\documentclass[runningheads]{llncs}
\usepackage[T1]{fontenc}
\usepackage{graphicx}
\usepackage{amsmath}
\usepackage{booktabs}
\usepackage{array}
\usepackage{cancel}
\usepackage{tabularx}
\usepackage{amsfonts}
\usepackage{amsfonts}
\usepackage{esvect}
\begin{document}
\title{Evaluating Generative Vehicle Trajectory Models for Traffic Intersection Dynamics}
\titlerunning{E.G.V.T.I.D}
%
\author{Yash Ranjan\thanks{Corresponding Author}
        \and Rahul Sengupta
        \and Anand Rangarajan
        \and Sanjay Ranka}
\authorrunning{Y. Ranjan et al.}
%
\institute{ University of Florida, Gainesville, FL, USA \\
            \email{\{yashranjan, rahulseng, anandr, sranka\}@ufl.edu}
            }

\maketitle              
\begin{abstract}
Traffic Intersections are vital to urban road networks as they regulate the movement of people and goods. However, they are regions of conflicting trajectories and are prone to accidents. Deep Generative models of traffic dynamics at signalized intersections can greatly help traffic authorities better understand the efficiency and safety aspects. At present, models are evaluated on computational metrics that primarily look at trajectory reconstruction errors. They are not evaluated online in a `live' microsimulation scenario. Further, these metrics do not adequately consider traffic engineering-specific concerns such as red-light violations, unallowed stoppage, etc.  In this work, we provide a comprehensive analytics tool to train, run, and evaluate models with metrics that give better insights into model performance from a traffic engineering point of view. We train a state-of-the-art multi-vehicle trajectory forecasting model on a large dataset collected by running a calibrated scenario of a real-world urban intersection. We then evaluate the performance of the prediction models, online in a microsimulator, under unseen traffic conditions. We show that despite using ideally-behaved trajectories as input, and achieving low trajectory reconstruction errors, the generated trajectories show behaviors that break traffic rules. We introduce new metrics to evaluate such undesired behaviors and present our results. 

\end{abstract}

\section{Introduction}
Traffic intersections play a critical role in the functioning of urban road networks. They ensure the smooth flow of vehicles and pedestrians by ensuring their trajectories do not conflict, avoiding congestion and accidents. Hence, dynamics seen at signalized urban intersections are complex, as various road users (vehicles and pedestrians) must not only interact with each other, but they must also pay attention to the traffic signal state as well as road markings indicating lanes and crosswalks. Computational-modeling of such scenarios was usually done on a case-by-case basis using agent-based micro-simulators, which can be computationally-heavy.

Intelligent Transportation Systems (ITS) has revolutionized the way sensor data from urban intersections is acquired, processed, and utilized. Various sensor modalities such as inductive loop detectors, signal state recorders, video cameras, LiDAR, radar etc. provide comprehensive high-resolution data.

Advances in Cloud Storage technologies have provided scalable data storage for vast amounts of multimodal sensor data. High-performance computing hardware (such as GPUs/TPUs) has accelerated the development of Deep Learning models trained on large datasets. Hence, there has been a great deal of published literature on the application of Deep Learning models to ITS. 

One important area of application is the trajectory prediction of road users.
Usually, trajectory data is generally obtained by performing object-tracking on video/LiDAR data. This often requires large data processing resources as well as advanced vision-processing algorithms \cite{Gupta2021DeepLF}.


\subsection{Related Work}

There are many works that deal with trajectory prediction for road users. Important among them include Trajectron (and its successor Trajectron++) \cite{traj++}, which represents a traffic scene as a directed spatio-temporal graph and produces dynamically-feasible trajectories using a graph-structured recurrent model. Another important model is TNT (target-driven trajectory) \cite{tnt} which consists of three stages that are trained end-to-end:  the first predicts an agent’s potential target states, the second generates trajectory state sequences conditioned on targets and the final stage estimates trajectory likelihoods and selects a final compact set of trajectory predictions. Surveys such as \cite{huang2023multimodal}, \cite{surveytraj}, and \cite{bharilya2023machinelearningautonomousvehicles} discuss various methods for vehicle trajectory prediction. However, they focus on autonomous vehicles and do not specifically deal with the modeling of intersection behaviors.

\cite{hagenus2024surveyrobustnesstrajectoryprediction} surveys commonly used performance metrics and robustness strategies for vehicle trajectory prediction. \cite{huang2023multimodaltrajectorypredictionsurvey} discusses commonly-used metrics and their drawbacks in the context of multimodal trajectory prediction. \cite{metrics1} introduces new metrics for trajectory prediction apart from commonly-used distance-based metrics. However, it does not focus on intersection-related behaviors, especially the effect of traffic signals and compliance thereof.

\subsection{Our Contribution}
While there is a good deal of relevant literature on deep generative models for trajectory prediction, there is a lack of metrics focusing on intersection dynamics in a multi-vehicle scenario. Present metrics are written from the point-of-view of autonomous vehicles, where the prediction of road users' trajectories revolves around the perception (via sensors) of the ego vehicle. This is not helpful to road traffic authorities who study the intersection dynamics as a whole, with trajectory data from intersection-mounted video cameras, LiDAR, DSRC/CV2X, drone-based cameras etc. These provide a birds-eye view of multiple vehicles at the intersection and their movements. 

Further, the evaluation of these models is generally done using a pre-recorded test set. It is not possible to generate new arrival patterns or change other parameters, such as signal timing plan.

Given the critical importance of signalized intersections to road traffic authorities, our work makes the following contributions:

\begin{enumerate}
    \item We develop an end-to-end framework for evaluating and benchmarking Deep Generative models for the task of multi-vehicle trajectory prediction at a signalized intersection. Given the trajectory prediction model architecture code, the system will train the model on the testbed intersection and vehicle flows, then evaluate the trained model and report its performance.
    \item We introduce new evaluation metrics that address traffic-specific concerns such as traffic signal compliance, unnecessary stoppage, and braking behavior. These are novel metrics not usually seen in published trajectory prediction literature.
    \item We include intersection-based vehicle position encoding and signal timing information in our trajectory prediction model, which is not usually seen in published literature.
    \item We test the models using a simulation-in-loop real-world signalized urban intersection with calibrated vehicle flows.
\end{enumerate}

\section{Trajectory Prediction Model for Signalized Intersection Scenario}

Our problem requires us to model the spatial-temporal relationship between an object’s history with the state of the objects around it to predict the future position. The model used is largely based on \cite{10588424} with key changes to incorporate intersection-specific details.

We discuss the different components of the model and how we incorporate the external factors which are the signal, intersection geometry, and the object position.

\begin{figure}
\centering
\includegraphics[width=0.80\textwidth]{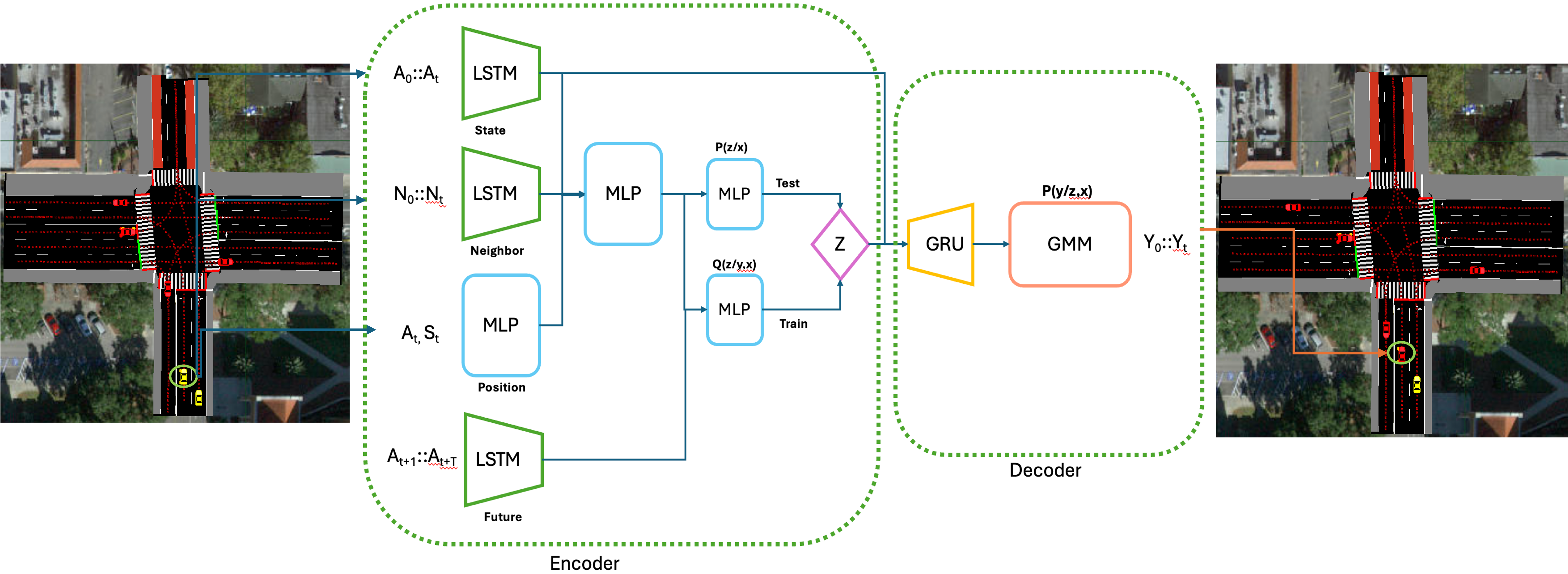}
\caption{The model architecture. The data comes from the simulator and is passed through the model to get the distribution over the future state. The mode of the distribution is sampled and sent back to the simulator, which visualizes it.}
\label{fig:model_architecture}
\vspace{-3mm}
\end{figure}

\subsection{Trajectory Prediction}

The objective of the trajectory prediction model is to generate plausible trajectory distributions for different agents (such as vehicles and pedestrians) that interact in a scene. This time-dependent number of agents is denoted by $n(t)$, representing the interacting agents $A_1, \ldots, A_{n(t)}$. In this work, we focus only on vehicles.

Each agent considers its own motion history, relevant traffic signals, potential interactions with neighboring agents, and its position relative to `static points' (described later). The historical data for agent $A_i$ at time $t$ is expressed as $X^{t-H+1,\ldots, t}_i$, where $H$ represents the number of history steps. The traffic signal for the route class $c$ of agent $A_i$ is denoted as $s^t_c$, its neighboring agents as $N^t_i$, and relative position $R^t_i$. The model outputs a probability distribution over the future trajectory $Y^{t+1, \ldots, t+T}_i$ for the next $T$ steps, formulated as $p(Y^{t+1, \ldots, t+T}_i | X^{t-H+1,\ldots, t}_i, N^t_i, R^t_i, s^t_c)$.

Here \( X \) consists of rectangular coordinates \( (x, y) \), velocity \( (\dot{x}, \dot{y}) \), acceleration \( (\ddot{x}, \ddot{y}) \) and Here \( Y \) consists of future rectangular coordinates \( (x, y) \).

\subsection{Agent History Modeling}

The agent history is modeled using an LSTM \cite{10.1162/neco.1997.9.8.1735} that captures the sequential relationship of the object's past. We pass the object's position, velocity, acceleration, and one-hot-encoded signal for the $H$ time stamp as input to the LSTM to generate the vector representation of the agent's state. The signal is the traffic light visible to the vehicle in its lane. Adding the signal in the object state helps us capture the transitional effect of the signal on the vehicle's trajectory. We denote the final encoded agent history as ${h}_i^{t}$.
\begin{equation}
    h^t_i = LSTM \left( h^{t-1}_i, \mathbf{x}^t_i, W \right).
\label{eq:hist_enc}
\end{equation}
During the training phase, we incorporate a Node Future Encoder (NFE) to capture a node’s true future trajectory \cite{ivanovic2019trajectronprobabilisticmultiagenttrajectory}. This encoder is based on a bi-directional LSTM network with 32 hidden dimensions, producing outputs denoted as \( h_{t+}^{i,\text{node}} \).

\subsection{Agent Interaction Modeling}
The agent-agent interactions in the scene are represented as a directed graph where the agents are the nodes and a directed edge is created between an agent and its neighbor if the neighbor is within the attention radius of the agent. This method is inspired from the Trajectron \cite{ivanovic2019trajectronprobabilisticmultiagenttrajectory} model.

At any given timestep, the states of the connected neighbors of an object type (VEHICLE or PEDESTRIAN) will be added and concatenated with the vehicle's state. The addition of neighbors' states is preferred instead of concatenation or taking average as it can handle a variable number of neighbors without losing count information \cite{jain2016structuralrnndeeplearningspatiotemporal}. This is passed through an LSTM to capture the changing relation of the vehicle with the objects in its attention radius.
\begin{equation}
    e^t_{i,k} = \left[ \mathbf{x}^t_i, \sum_{j \in N_k(i)} \mathbf{x}^t_j \right],
\label{edge_state_concat}
\end{equation}
\begin{equation}
    h^t_{i,k} = LSTM \left( h^{t-1}_{i,k}, e^t_{i,k}; W \right).
\label{edge_encoder}
\end{equation}

We use additive attention mechanism to combine the embedding vectors and get a single neighbor influence embedding.
\begin{equation}
    s^t_{ik} = \mathbf{v}^T_{C_i} \tanh \left( W_{1,C_i} h^t_{i,k} + W_{2,C_i} h^t_{i,\text{node}} \right),
\label{add_attention}
\end{equation}
\begin{equation}
    a^t_i = \text{softmax} \left( [s^t_{i1}, \dots, s^t_{iK}] \right) \in \mathbb{R}^K,
\label{att_soft_max}
\end{equation}
\begin{equation}
    h^t_{i,\text{edges}} = \sum_{k=1}^{K} a^t_{ik} \odot h^t_{i,k}.
\label{att_weighted_sim}
\end{equation}
\subsection{Node Position Embedding}
\begin{figure}[!htbp]
\centering
\includegraphics[width=0.60\textwidth]{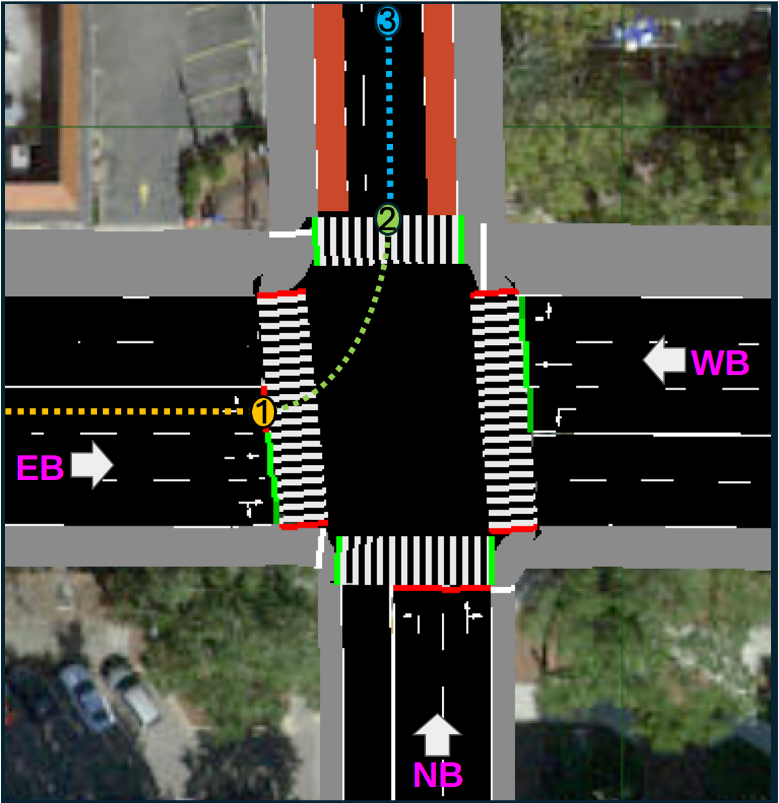}
\caption{The trajectory is divided into three regions. First is the incoming region, second is the in-between region, and third is the outgoing region. Each region has an end point which acts as the static point. The position embedding of the vehicle is found out relative to the static point based on the region the vehicle is}
\label{fig:regions}
\vspace{-3mm}
\end{figure}

As the trajectory of the vehicle depends on the position of the vehicle at the intersection, we came up with a simple but effective way to embed the position of the vehicle. We divided the trajectories of vehicles into three regions as shown in Figure \ref{fig:regions}. Each region has an endpoint 
that acts as a static point and the positional embedding of the vehicle is found relative to the static point based on the region that the vehicle belongs to in its trajectory. The idea behind this is that as a driver when we are approaching an intersection, we keep track of the end of the region in our mind. So we want to encode this distance in a vector. To get the positional embedding, we pass the current coordinate of the object, the coordinate of the static point based on the region that the vehicle belongs to, and the Euclidean distance between them. We also pass the one-hot encoding of the region as well as the one-hot encoding of the trajectory (or cluster) that the vehicle belongs to \cite{10588424}.
\begin{equation}
    p^t_i = MLP(x^t_i, x_{st}, d^t_i, c_i, r^t_i).
\label{eq:pos_enc}
\end{equation}
Here $x_{st}$ is the coordinate of the next static point, $d^t_i$ is the distance between the object and the static point, $c_i$ is the cluster that the vehicle belongs to and $r^t_i$ is the region of the trajectory that the vehicle is in.

\subsection{Conditional Variational AutoEncoder}
We combine the three encoded vectors by concatenating them and passing them through a Multi-Layer Perceptron. This acts as the embedded information on which the model is conditioned to predict trajectories.
\begin{equation}
    h^t_{i,enc} = MLP(h^t_i, h^t_{i,k}, p^t_i).
\end{equation}
We followed Trajectron++\cite{salzmann2021trajectrondynamicallyfeasibletrajectoryforecasting} to model the probability distribution over the trajectories \( p(y \mid x) \) using the CVAE framework and introduced the discrete latent variable $z$ to get the equation:
\begin{equation}
    p(y|x) = \sum_{z \in Z} p_\theta( y| x, z) p_\phi (z| x)),  
\end{equation}
where discrete latent variable $z$ models different modes for predicting future trajectories $y$ given the input condition $x$. $p_\theta( {y}| {x}, z)$ and $ p_\phi (z| {x})$ are sub-networks with learnable parameters $\theta$ and $\phi$. Instead of producing a deterministic trajectory prediction, the CVAE \cite{Pagnoni2018ConditionalVA} model is able to predict a distribution of plausible future trajectories, accounting for the variability of vehicle and pedestrian behavior. For example, when drivers interact with a potentially conflicting pedestrian, some may opt to yield, while others may decide to accelerate and pass first. Under different situations, the probability of each behavior varies. We set $|Z| = 25$ in our experiments. 

\subsection{Decoder Network}
During training the model, we get the latent variable probability from the output of the network \( q({z}|{x},y) \). This is nothing but the mixture weights for the Gaussian Mixture Model that we use to model the future trajectories. During inference we sample from \( p(z| x)\). We then send the sampled $z$ and $h^t_{i,enc}$ to the decoder. We are using a GRU network with hidden dimensions of 128. The decoded vectors are then passed through mixture density network \cite{370fbeadb5584ba9ab2938431fc4f140} to get the mean and covariance of the Gaussian for each $z$ component for the next time-step. We do this for the next many time-steps.

For model training, we employ the CVAE's objective function, as described in \cite{zhao2018infovaeinformationmaximizingvariational} with the Adam optimizer:
\begin{equation}
    \max_{\phi, \theta, \psi} \sum_{i=1}^{N} \mathbb{E}_{z \sim q_\phi(z \mid x_i, y_i)} 
    \left[ \log p_\psi(y_i \mid x_i, z) \right]
    - \beta D_{KL}(q_\phi(z \mid x_i, y_i) \parallel p_\theta(z \mid x_i)).
\label{eq:loss_func}
\end{equation}
During inference, we sample the probability distribution $y \sim p_\theta ({y} | {x}, z)$ to obtain the future trajectories. The sampling can be performed by either finding the mixture weights of the GMM from the model \( p(z\mid x)\)  and then sampling the most likely output of the distribution $y\sim p_\theta ({y} | {x}, z)$ or sampling the individual Gaussians to obtain the trajectories for the different modes.

\section{Intersection-aware Evaluation Metrics}

We summarize commonly-used metrics in Table \ref{tab:metrics}, \cite{huang2023multimodal}, \cite{surveytraj}, \cite{bharilya2023machinelearningautonomousvehicles}. We then describe new intersection-aware evaluation metrics for intersection-focused scenarios.

\begin{table}[htbp]
    \centering
    \caption{Summary of Evaluation Metrics for Trajectory Prediction Models.}
    \begin{tabular}{>{\raggedright\arraybackslash}p{3cm} >{\raggedright\arraybackslash}p{5cm} >{\raggedright\arraybackslash}p{4cm}}
        \toprule
         Metric & Expression & Description \\
        \midrule
        Average Displacement Error (ADE), minADE & 
        \[
        \frac{1}{T} \sum_{t=1}^{T} \sqrt{(x_t - \hat{x}_t)^2 + (y_t - \hat{y}_t)^2}
        \]
        & Mean Euclidean distance between the predicted trajectory locations and the corresponding ground truth locations over all time steps.  \\
        \addlinespace
        Final Displacement Error (FDE), minFDE & 
        \[
        \sqrt{(x_T - \hat{x}_T)^2 + (y_T - \hat{y}_T)^2}
        \]
        & Euclidean distance between the predicted final position and the true final position of the trajectory. \\
        \addlinespace
        KDE Negative Log-likelihood (KDE-NLL) &
        \[
        - \mathbb{E}_{i,t \in \tau} \log P(Y_t^i | \text{KDE} (\hat{Y}_t^{i,K}))
        \]
        & Evaluates the mean log-likelihood of the ground truth trajectory for each future timestep using Kernel Density Estimation for handling multi-modal trajectory predictions. \\
        \midrule
    \end{tabular}

    \label{tab:metrics}
\end{table}

\subsection{Red Light Violation}
This metric tracks the number of times a vehicle breaks the red light signal. 
The traffic phases for a 4-approach intersection are usually based on NEMA 8-phase numbering\footnote{https://ops.fhwa.dot.gov/publications/fhwahop08024/chapter4.htm}. At an approach, a vehicle can take the following turn-movements:

\begin{enumerate}
    \item Right-turn
    \item Through (go straight)
    \item Left-turn (and a U-turn)
\end{enumerate}

It is important to note that taking a right-turn during a red light is often allowed at many intersections (after yielding). However, in our experiments, this sort of right turn on red light is not allowed.

\subsection{Mid-intersection Stoppage}
This occurs when the vehicle unexpectedly stops in the middle of the intersection (i.e. in the area enclosed by the 4 crosswalks) and refuses to move. In some cases, this may be valid, as a left-turning vehicle may be waiting (yielding) for an on-coming vehicle to pass or a pedestrian to cross. Since in our experiments, there are no such conflicting movements, there should be no such stoppage. If a vehicle remains stopped at such a location for an extended period of time (5 mph/8 kmph or less, for 2 seconds at least), we flag that as an invalid stoppage. 

\subsection{Pre-Stopbar Stoppage}

This occurs when a vehicle does not respond properly to a signal transition from Red to Green light. The vehicle remains stopped or advances too slowly despite no vehicles ahead of it blocking its path. When the light is green, we take a queue discharge rate of 0.5 m/s, which would translate to the discharge of 1 vehicle (effective length including headway) of 7 meters, in 14 seconds \cite{TQAM}. This is a low-end estimate; hence the vehicle should cover its distance to the stopbar faster than this during the green time it sees. If it does not, it could be a cause for concern and require further investigation. It was noticed that during the experiments, vehicles often encounter a queue, slow down/stop, and then are unable to start fast enough as the queue discharges. Thus, they block the lane despite having sufficient distance ahead of them to move.

\subsection{Time-To-Collision (TTC) Encounters}
TTC is the time it would take for two vehicles to collide, assuming their current speeds and paths remain unchanged. It is calculated based on the relative distance between the vehicles and their relative speed. TTC is a well-known metric in traffic safety analysis but is rarely used in trajectory models \footnote{https://www.fhwa.dot.gov/publications/research/safety/03050/02.cfm}.

Let:
\begin{itemize}
    \item $d$ be the distance between two vehicles.
    \item $v_{\text{rel}}$ be the relative speed between the two vehicles.
\end{itemize}

Then, the Time to Collision (TTC) is given by:

\begin{equation}
    \text{TTC} = \frac{d}{v_{\text{rel}}},
\label{ttc_eqn}
\end{equation}

where:
\begin{itemize}
    \item $d$ is the distance between the two vehicles.
    \item $v_{\text{rel}} = |v_1 - v_2|$ is the absolute value of the relative speed between the vehicles, where $v_1$ and $v_2$ are the speeds of the two vehicles.
\end{itemize}

A high TTC indicates a safer situation, as there is more time before a potential collision. We use SUMO simulator's in-built TTC calculation module\footnote{https://sumo.dlr.de/docs/Simulation/Output/SSM\_Device.html} to extract the TTC values. SUMO reports `conflict' events where the TTC is low enough to be of concern. We use 4 seconds \cite{ttc1} as the threshold, below which TTC values are of concern. Each `conflict' event can be considered a concerning encounter. Different encounters may include the same vehicles but at different points in their trajectories.

\section{Results and Discussion}

\begin{table}[!htbp]
\setlength{\tabcolsep}{6pt}
    \centering
    \scriptsize
    \caption{Table of Results for Improved Model. Note that different TTC Encounters can include the same vehicles but at different points in their trajectories.}
\begin{tabular}{lccccc}
\toprule
 & Total & Red Light & Mid-Intersection & Pre-Stopbar & TTC \\
 & Count & Violation & Stoppage & Stoppage & Encounters \\
Trajectory Cluster &  &  &  &  &  \\
\midrule
L on EBL & 102 & 0 (0.0\%) & 0 (0.0\%) & 69 (67.6\%) & 139 \\
L on NBL & 133 & 0 (0.0\%) & 1 (0.8\%) & 47 (35.3\%) & 207 \\
L on WBL & 127 & 25 (19.7\%) & 0 (0.0\%) & 1 (0.8\%) & 109 \\
R on EBTR & 99 & 0 (0.0\%) & 0 (0.0\%) & 0 (0.0\%) & 139 \\
R on NBTR & 115 & 0 (0.0\%) & 0 (0.0\%) & 20 (17.4\%) & 118 \\
R on WBTR & 106 & 69 (65.1\%) & 0 (0.0\%) & 3 (2.8\%) & 57 \\
T on EBT & 57 & 0 (0.0\%) & 0 (0.0\%) & 35 (61.4\%) & 36 \\
T on EBTR & 61 & 0 (0.0\%) & 0 (0.0\%) & 7 (11.5\%) & 85 \\
T on NBTR & 111 & 0 (0.0\%) & 0 (0.0\%) & 15 (13.5\%) & 123 \\
T on WBT & 36 & 13 (36.1\%) & 0 (0.0\%) & 0 (0.0\%) & 12 \\
T on WBTR & 79 & 45 (57.0\%) & 0 (0.0\%) & 0 (0.0\%) & 22 \\
\midrule
Total & 1026 & 152 (14.8\%)& 1 (0.0\%)& 197 (19.2\%)& 1047 \\
\midrule
\end{tabular}

    \label{tab:resultsi}
\end{table}

In order to evaluate our metrics under `live' simulation settings, we first train our model on a dataset obtained by simulating a traffic intersection in SUMO\footnote{\url{sumo.dlr.de/docs/SUMO\_at\_a\_Glance.html}} microscopic traffic simulator. The dataset was based on 36 hours of simulated data, based on a real-world intersection, calibrated with real-world traffic flows seen during the day. The average vehicle flow is between 900-1100 vehicles per hour. Signal cycle length was set at 90 seconds with actuated phases. The intersection simulated consists of 4 approaches as shown in Figure \ref{fig:regions}. Note that the approach on the upper portion is a 1-way street, where vehicles can only leave the intersection. Hence, only at the other 3 approaches i.e. North-bound (NB), East-bound (EB) and West-bound (WB) approaches can the vehicles approach the intersection, stop (if required) and then cross the intersection. There are multiple lanes on these approaches:

\begin{itemize}
    \item NB: 2 lanes, that allow Through/Right (TR) and only Left (L) turns.
    \item WB: 3 lanes, that allow Through/Right (TR), only Through (T) and only Left (L) turns.
    \item EB: 3 lanes, that allow Through/Right (TR), only Through (T) and only Left (L) turns.
\end{itemize}

While a vehicle on `only Through' lane can go only straight, one on the `Through/Right' may go straight or take a right turn. This leads to 11 possible trajectory clusters, based on which direction and lane the vehicle approaches the intersection, and then which turn the vehicle makes. For example, a trajectory cluster could be of a West-bound vehicle on the `Through/Right' lane, deciding to take a right turn: R on WBTR. A vehicle can only belong to one cluster.

This intersection is a real-world intersection in an urban area of North American city, bordering a large university campus. The simulation model has been calibrated based on flow rates, speeds etc.

Once the generative model has been trained to convergence (with reasonable accuracy based on pre-existing metrics in Table \ref{tab:metrics}), we freeze the weights. The PyTorch\footnote{pytorch.org} model is coupled with SUMO simulator via TRACI interface. The first 2 seconds of ground-truth trajectory is provided for each vehicle as it enters the simulation, as the initial condition to start the model prediction. Then the model predicts the rest of the trajectory till the vehicle exits the simulation \cite{10588424}. We use 4000 seconds (1 hour 6 minutes) of simulation data, for 1026 vehicles in total. Multiple vehicles may enter the scene at any given time, and the model must unroll all the trajectories simultaneously. Once the predicted trajectories are unrolled, simulation logs from SUMO are collected. MovingPandas library \cite{graser_movingpandas_2019} was used for calculating the metrics.

\begin{table}[htbp]
\setlength{\tabcolsep}{6pt}
    \centering
    \scriptsize
    \caption{Table of Results for Baseline Model.  Note that different TTC Encounters can include the same vehicles but at different points in their trajectories.}
\begin{tabular}{lccccc}
\toprule
 & Total & Red Light & Mid-Intersection & Pre-Stopbar & TTC \\
 & Count & Violation & Stoppage & Stoppage & Encounters \\
Trajectory Cluster &  &  &  &  &  \\
\midrule
L on EBL & 102 & 37 (36.3\%) & 0 (0.0\%) & 13 (12.7\%) & 55 \\
L on NBL & 133 & 3 (2.3\%) & 1 (0.8\%) & 66 (49.6\%) & 202 \\
L on WBL & 127 & 13 (10.2\%) & 35 (27.6\%) & 3 (2.4\%) & 118 \\
R on EBTR & 99 & 49 (49.5\%) & 0 (0.0\%) & 4 (4.0\%) & 37 \\
R on NBTR & 115 & 18 (15.7\%) & 1 (0.9\%) & 27 (23.5\%) & 175 \\
R on WBTR & 106 & 0 (0.0\%) & 0 (0.0\%) & 29 (27.4\%) & 89 \\
T on EBT & 27 & 15 (55.6\%) & 0 (0.0\%) & 0 (0.0\%) & 9 \\
T on EBTR & 91 & 8 (8.8\%) & 0 (0.0\%) & 4 (4.4\%) & 65 \\
T on NBTR & 111 & 4 (3.6\%) & 2 (1.8\%) & 43 (38.7\%) & 208 \\
T on WBT & 61 & 9 (14.8\%) & 8 (13.1\%) & 13 (21.3\%) & 49 \\
T on WBTR & 54 & 0 (0.0\%) & 0 (0.0\%) & 12 (22.2\%) & 34 \\
\midrule
Total & 1026 & 156 (15.2\%)& 47 (4.5\%)& 214 (20.8\%)& 1041 \\
\midrule
\end{tabular}

    \label{tab:resultsb}
\end{table}

We train our novel `Improved Model' in which we have incorporated an intersection-based vehicle position encoder (generalizable across any intersection geometry) as well as the traffic signal state, while predicting future trajectories. The `Baseline' model (based on \cite{10588424}) only incorporates signal timing information. The original Trajectron++ paper \cite{traj++} incorporates neither, and is not included in the analysis.

We tabulate the results for the Improved Model in Table \ref{tab:resultsi} and for the Baseline Model in \ref{tab:resultsb}, stratified by trajectory clusters. We see that both these models show significant Red Light Violation (around 15 \%) and Pre-Stopbar Stoppage (around 20 \%). However, the `Improved Model' virtually eliminates all Mid-Intersection Stoppage violations. These are incredibly dangerous, as stopping in the middle of the intersection can lead to extreme accidents, due to the high speeds involved. The Time-To-Collision Encounters of concern remain significantly high. These phenomena would not have been captured by the existing metrics (Table \ref{tab:metrics}). With these new metrics, we are able to detect and characterize previously unstudied issues.

\section{Conclusion}

In this work, we trained two models on vehicle trajectories near an intersection and then evaluated them in an unseen `live' simulation scenario. We calculated the intersection-aware metrics: Red Light Violation, Mid-Intersection Stoppage, Pre-Stopbar Stoppage, and Time-To-Collision Encounters. The `Improved Model' included a novel intersection-based vehicle position encoding, which virtually eliminated Mid-Intersection Stoppage events. However, significant issues remain; they pose a serious public hazard and need to be fixed. We hope that the findings in this work, along with the metrics presented, will help researchers and policy-makers deploy safer and more robust multi-vehicle trajectory models.

%
%
%
\bibliographystyle{splncs04}
 \bibliography{pakdd}

\begin{thebibliography}{10}
\providecommand{\url}[1]{\texttt{#1}}
\providecommand{\urlprefix}{URL }
\providecommand{\doi}[1]{https://doi.org/#1}

\bibitem{bharilya2023machinelearningautonomousvehicles}
Bharilya, V., Kumar, N.: Machine learning for autonomous vehicle's trajectory prediction: A comprehensive survey, challenges, and future research directions (2023), \url{https://arxiv.org/abs/2307.07527}

\bibitem{370fbeadb5584ba9ab2938431fc4f140}
Bishop, C.: Mixture density networks. Workingpaper, Aston University (1994)

\bibitem{metrics1}
Chen, C., Pourkeshavarz, M., Rasouli, A.: Criteria: a new benchmarking paradigm for evaluating trajectory prediction models for autonomous driving. In: 2024 IEEE International Conference on Robotics and Automation (ICRA). pp. 8265--8271 (2024). \doi{10.1109/ICRA57147.2024.10610911}

\bibitem{graser_movingpandas_2019}
Graser, A.: {MovingPandas}: {Efficient} {Structures} for {Movement} {Data} in {Python}. GI\_Forum ‒ Journal of Geographic Information Science  \textbf{7}(1),  54--68 (2019). \doi{10.1553/giscience2019_01_s54}, \url{https://hw.oeaw.ac.at?arp=0x003aba2b}

\bibitem{Gupta2021DeepLF}
Gupta, A., Anpalagan, A., Guan, L., Khwaja, A.S.: Deep learning for object detection and scene perception in self-driving cars: Survey, challenges, and open issues. Array  \textbf{10},  100057 (2021), \url{https://api.semanticscholar.org/CorpusID:233562996}

\bibitem{hagenus2024surveyrobustnesstrajectoryprediction}
Hagenus, J., Mathiesen, F.B., Schumann, J.F., Zgonnikov, A.: A survey on robustness in trajectory prediction for autonomous vehicles (2024), \url{https://arxiv.org/abs/2402.01397}

\bibitem{10.1162/neco.1997.9.8.1735}
Hochreiter, S., Schmidhuber, J.: Long short-term memory. Neural Comput.  \textbf{9}(8),  1735–1780 (Nov 1997). \doi{10.1162/neco.1997.9.8.1735}, \url{https://doi.org/10.1162/neco.1997.9.8.1735}

\bibitem{huang2023multimodal}
Huang, R., Xue, H., Pagnucco, M., Salim, F., Song, Y.: Multimodal trajectory prediction: A survey. arXiv preprint arXiv:2302.10463  (2023)

\bibitem{huang2023multimodaltrajectorypredictionsurvey}
Huang, R., Xue, H., Pagnucco, M., Salim, F., Song, Y.: Multimodal trajectory prediction: A survey (2023), \url{https://arxiv.org/abs/2302.10463}

\bibitem{surveytraj}
Huang, Y., Du, J., Yang, Z., Zhou, Z., Zhang, L., Chen, H.: A survey on trajectory-prediction methods for autonomous driving. IEEE Transactions on Intelligent Vehicles  \textbf{7}(3),  652--674 (2022). \doi{10.1109/TIV.2022.3167103}

\bibitem{ivanovic2019trajectronprobabilisticmultiagenttrajectory}
Ivanovic, B., Pavone, M.: The trajectron: Probabilistic multi-agent trajectory modeling with dynamic spatiotemporal graphs (2019), \url{https://arxiv.org/abs/1810.05993}

\bibitem{jain2016structuralrnndeeplearningspatiotemporal}
Jain, A., Zamir, A.R., Savarese, S., Saxena, A.: Structural-rnn: Deep learning on spatio-temporal graphs (2016), \url{https://arxiv.org/abs/1511.05298}

\bibitem{Pagnoni2018ConditionalVA}
Pagnoni, A., Liu, K., Li, S.: Conditional variational autoencoder for neural machine translation. ArXiv  \textbf{abs/1812.04405} (2018), \url{https://api.semanticscholar.org/CorpusID:54555710}

\bibitem{ttc1}
Ramezani-Khansari, E., Moogehi, S., Moghadas~Nejad, F.: Comparing time to collision and time headway as safety criteria. Pamukkale University Journal of Engineering Sciences  \textbf{27} (06 2020). \doi{10.5505/pajes.2020.79837}

\bibitem{traj++}
Salzmann, T., Ivanovic, B., Chakravarty, P., Pavone, M.: Trajectron++: Dynamically-feasible trajectory forecasting with heterogeneous data (2021), \url{https://arxiv.org/abs/2001.03093}

\bibitem{salzmann2021trajectrondynamicallyfeasibletrajectoryforecasting}
Salzmann, T., Ivanovic, B., Chakravarty, P., Pavone, M.: Trajectron++: Dynamically-feasible trajectory forecasting with heterogeneous data (2021), \url{https://arxiv.org/abs/2001.03093}

\bibitem{TQAM}
Sengupta, R., Karnati, Y., Rangarajan, A., Ranka, S.: Tqam: Temporal attention for cycle-wise queue length estimation using high-resolution loop detector data. In: 2021 IEEE International Intelligent Transportation Systems Conference (ITSC). pp. 3313--3320 (2021). \doi{10.1109/ITSC48978.2021.9564900}

\bibitem{10588424}
Wu, A., Ranjan, Y., Sengupta, R., Rangarajan, A., Ranka, S.: A data-driven approach for probabilistic traffic prediction and simulation at signalized intersections. In: 2024 IEEE Intelligent Vehicles Symposium (IV). pp. 3092--3099 (2024). \doi{10.1109/IV55156.2024.10588424}

\bibitem{tnt}
Zhao, H., Gao, J., Lan, T., Sun, C., Sapp, B., Varadarajan, B., Shen, Y., Shen, Y., Chai, Y., Schmid, C., Li, C., Anguelov, D.: Tnt: Target-driven trajectory prediction (2020), \url{https://arxiv.org/abs/2008.08294}

\bibitem{zhao2018infovaeinformationmaximizingvariational}
Zhao, S., Song, J., Ermon, S.: Infovae: Information maximizing variational autoencoders (2018), \url{https://arxiv.org/abs/1706.02262}

\end{thebibliography}

\end{document}